\documentclass[11pt,a4paper]{article}

\usepackage[hyperref]{emnlp-ijcnlp-2019}
\usepackage{times}
\usepackage{latexsym}
\usepackage{amsmath}
\usepackage{amsfonts}
\usepackage{booktabs}
\usepackage{colortbl}
\usepackage{verbatim}

\usepackage{url}
\usepackage{graphicx}

\usepackage{algorithm}
\usepackage[noend]{algpseudocode}
\usepackage{textcomp}

\aclfinalcopy 

\DeclareMathOperator*{\argmax}{arg\,max}

\newcommand{\ccell}{\begin{tabular}[t]{@{}c@{}}\end{tabular}}

\newcommand{\bottleEx}{{\usefont{T1}{pzc}{m}{n} BottleSum$^{Ex}$}}

\newcommand{\bottleSelf}{{\usefont{T1}{pzc}{m}{n} BottleSum$^{Self}$}}
\newcommand{\bottleSum}{{\usefont{T1}{pzc}{m}{n} BottleSum}}

\newcommand{\reconEx}{{ Recon$^{Ex}$}}

\newcommand\peter[1]{{\color{purple}\{\textit{#1}\}$_{peter}$}}

\title{\bottleSum: Unsupervised and Self-supervised Sentence Summarization \\using the Information Bottleneck Principle}

\author{Peter West \\
  Paul G. Allen School of Computer Science & Engineering, University of Washington \\
  {\tt pawest@cs.washington.edu} \\\And
  Ati Holtzman \\
  Paul G. Allen School of Computer Science & Engineering, University of Washington \\
  {\tt ahai@cs.washington.edu} \\\And
  Jan Buys \\
  Paul G. Allen School of Computer Science & Engineering, University of Washington \\
  {\tt jbuys@cs.washington.edu} \\\And
  Yejin Choi \\
  Paul G. Allen School of Computer Science & Engineering, University of Washington \\
  Allen Institute for Artificial Intelligence\\
  {\tt yejin@cs.washington.edu} \\}
  
\author{Peter West\textsuperscript{1} \hspace{1cm} Ari Holtzman\textsuperscript{1,2} \hspace{1cm} Jan Buys\textsuperscript{1} \hspace{1cm} Yejin Choi\textsuperscript{1,2}\\
  \textsuperscript{1}Paul G. Allen School of Computer Science \& Engineering, University of Washington\\
  \textsuperscript{2}Allen Institute for Artificial Intelligence\\
  {\tt \{pawest, ahai, jbuys, yejin\}@cs.washington.edu}}

\date{}

\begin{document}
\maketitle
\begin{abstract}

The principle of the Information Bottleneck \cite{tishby99information} 
is to produce a summary of information $X$ optimized to predict some other relevant information $Y$. 
In this paper, we propose a novel approach to unsupervised sentence summarization by mapping the Information Bottleneck principle to a conditional language modelling objective: given a sentence, our approach seeks a compressed sentence that can best predict the \emph{next} sentence.  
Our iterative algorithm under the Information Bottleneck objective searches gradually shorter subsequences of the given sentence while maximizing the probability of the next sentence conditioned on the summary. Using only pretrained language models with no direct supervision, our approach can efficiently perform extractive sentence summarization over a large corpus. 

Building on our \emph{unsupervised extractive} summarization (\bottleEx{}), we then present a new approach to \emph{self-supervised abstractive} summarization (\bottleSelf{}), where a transformer-based language model is trained on the output summaries of our unsupervised method. 
Empirical results demonstrate that our extractive method outperforms other unsupervised models on multiple automatic metrics. 
In addition, we find that our self-supervised abstractive model outperforms unsupervised baselines (including our own) by human evaluation along multiple attributes. 
\end{abstract}

\section{Introduction} \label{sec:intro}

\begin{figure}[t!]
    \centering
    \includegraphics[width=215pt,trim={17 0 0cm 0},clip]{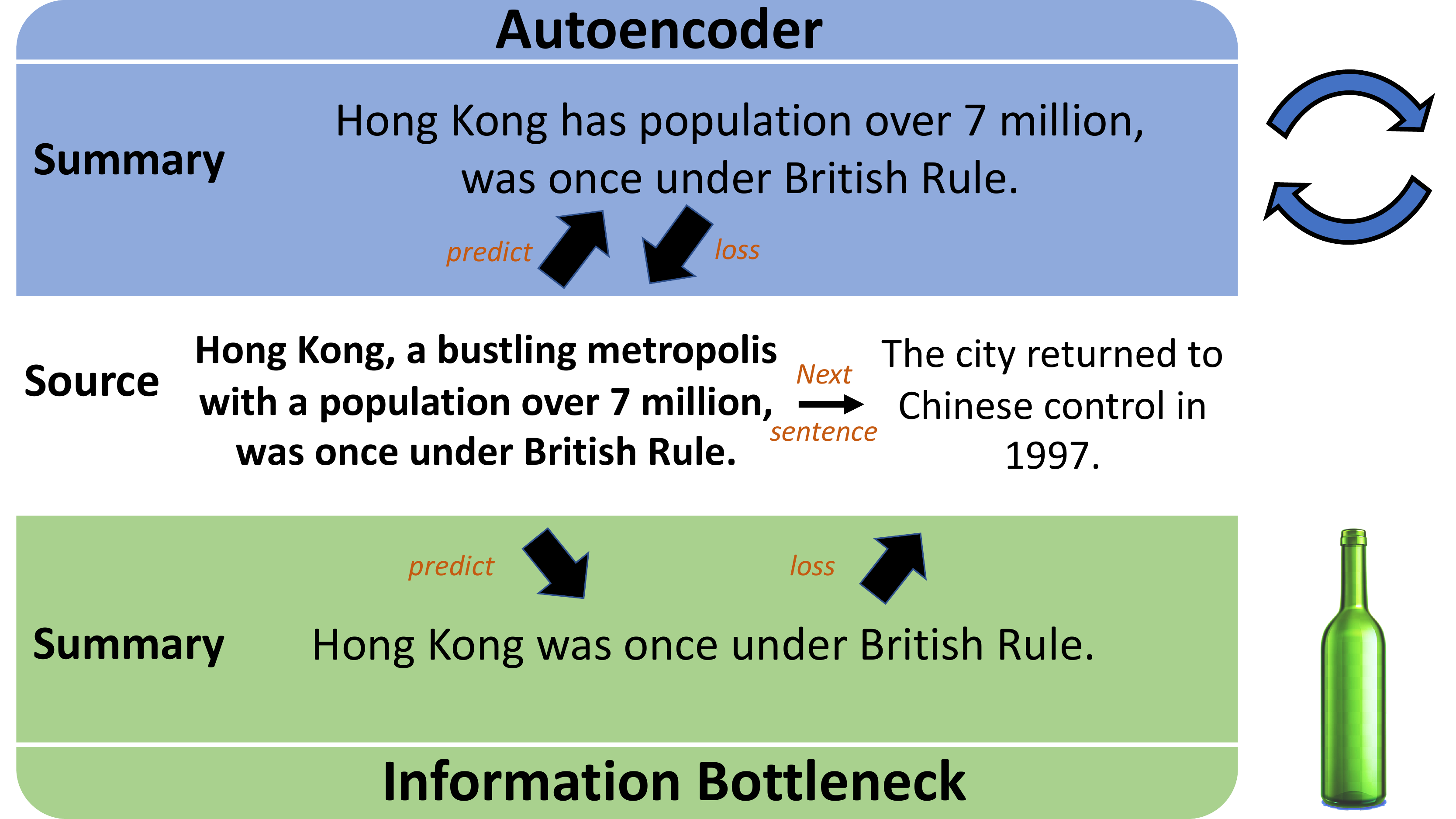}
    \caption{Example contrasting the Autoencoder (AE) and Information Bottleneck (IB) approaches to summarization. 
    While AE (top) preserves any detail that helps to reconstruct the original, such as population size in this example, IB (bottom) uses context to determine which information is relevant, which results in a more appropriate summary.
    }
    \label{fig:fig1}
\end{figure}

Recent approaches based on neural networks have brought significant advancements for both extractive and abstractive  summarization \cite{rush2015namas,nallapati2016abstractive}. However, their success relies on large-scale parallel corpora of input text and output summaries for direct supervision. For example, there are \texttildelow280,000 training instances in the CNN/Daily Mail dataset \cite{nallapati2016abstractive,hermann2015cnn},  and \texttildelow4,000,000 instances in the  sentence summarization dataset of \newcite{rush2015namas}. 
Because it is too costly to have humans write gold summaries at this scale, existing large-scale datasets are based on naturally occurring pairs of summary-like text paired with source text, for instance using news titles or highlights as summaries for news-text.
A major drawback to this approach is that these pairs must already exist in-domain, which is often not true.

The sample inefficiency of current neural approaches limits their impact across different tasks and domains, motivating the need for unsupervised or self-supervised alternatives \cite{Artetxe-2017-unsupnmt,LeCun, urgen1990making}.  
Further, for summarization in particular, the current paradigm requiring millions of supervision examples is almost counter-intuitive; after all, humans don't need to see a million summaries to know how to summarize, or what information to include. 

In this paper, we present  \bottleSum, consisting of a pair of novel approaches, \bottleEx{} and \bottleSelf{} for \emph{unsupervised extractive} and \emph{self-supervised abstractive} summarization, respectively. Core to our approach is the principle of the Information Bottleneck \cite{tishby99information}, producing a summary for information X optimized to predict some other relevant information Y. In particular, we map (conditional) language modeling objectives to the Information Bottleneck principle to guide the unsupervised model on what to keep and what to discard. 

The key intuition of our bottleneck-based summarization is that a good sentence summary contains information related to the broader context while discarding less significant details. Figure~\ref{fig:fig1} demonstrates this intuition. Given input sentence ``Hong Kong, a bustling metropolis with a population over 7 million, ...'', which is followed by the next sentence ``The city returned to Chinese control in 1997'', the information bottleneck would suggest that minute details such as the city's population being over 7 million are relatively less important to keep. In contrast, the continued discussion of the city's governance in the next sentence suggests its former British rule is important here.

This intuition contrasts with that of autoencoder-based approaches where the goal is to minimize the reconstruction loss of the input sentence when constructing the summary 
\cite{miao-blunsom-2016-language,wang-lee-2018-learning,fevry-phang-2018-unsupervised,SEQ3}. 
Under the reconstruction loss, minute but specific details such as the city's population being over 7 million will be difficult to discard from the summary, because they are useful for reconstruction.

Concretely, \bottleEx{} is an extractive and unsupervised sentence summarization method using the next sentence, a sample of nearby context, as guidance to \emph{relevance}, or what information to keep. We capture this with a conditional language modelling objective, allowing us to benefit from powerful deep neural language models that are pre-trained  over an extremely large-scale corpus. Under the Information Bottleneck objective, we present an iterative algorithm that searches gradually shorter subsequences of the source sentence while maximizing the probability of the next sentence conditioned on the summary. The benefit of this approach is that it requires no domain-specific supervision or fine-turning. 

Building on our unsupervised extractive summarization, we then present \bottleSelf{}, a new approach to self-supervised abstractive summarization. This method also uses a pretrained language model, but turns it into an abstractive summarizer by fine-tuning on the output summaries generated by \bottleEx{} paired with their original input sentences. 
The goal is to generalize the summaries generated by an extractive method by training a language model on them, which can then produce abstractive summaries as its generation is not constrained to be extractive. 

Together, \bottleEx{} and \bottleSelf{} are \bottleSum{ } methods for unsupervised sentence summarization. Empirical  results  demonstrate  that \bottleEx{}  outperforms other unsupervised methods on multiple automatic metrics, closely followed by \bottleSelf{}.   Furthermore, testing on a large unsupervised corpus, we find \bottleSelf{} outperforms unsupervised baselines (including our own \bottleEx{}) on human evaluation along multiple attributes.

\section{The Information Bottleneck Principle}\label{sec:IB}

Unsupervised summarization requires formulating an appropriate learning objective that can be optimized without supervision (example summaries). 
Recent work has treated unsupervised summarization as an autoencoding problem with a reconstruction loss \cite{miao-blunsom-2016-language,SEQ3}. The goal is then to produce a compressed summary from which the source sentence can be accurately predicted, i.e. to maximize:
\begin{equation} \label{eq:reconstructive}
 \mathbb{E}_{p(\tilde{s} | s)}\ \log p(s|\tilde{s}),
\end{equation}
where $s$ is the source sentence, $\tilde{s}$ is the generated summary and $p(\tilde{s} | s)$ the learned summarization model. The exact form of this loss may be more elaborate depending on the system, for example including an auxiliary language modeling loss, but the main aim is to produce a summary from which the source can be reconstructed.

The intuitive limitation of this approach is that it will always prefer to retain all informative content from the source. 
This goes against the fundamental goal of summarization, which crucially needs to forget all but the ``relevant'' information. It should be detrimental to keep tangential information, as illustrated by the example in Figure~\ref{fig:fig1}. As a result, autoencoding systems need to introduce additional loss terms to augment the reconstruction loss (e.g. length penalty, or the topic loss of \newcite{SEQ3}). 

The premise of our work is that the Information Bottleneck (IB) principle \cite{tishby99information} is a more natural fit for summarization. Unlike reconstruction loss, which requires augmentative terms to summarize, IB naturally incorporates a tradeoff between information selection and pruning. These approaches are compared directly in section \ref{sec:experiments}.

At its core, IB is concerned with the problem of maximal compression while defining a formal notion of information relevance. This is introduced with an external variable $Y$. 
The key is that $\tilde{S}$, the summary of source $S$, contains only information useful for predicting $Y$. This can be posed formally as learning a conditional distribution $p(\tilde{S}|S)$ minimizing:
\begin{equation} \label{eq:IB}
    I(\tilde{S};S) - \beta I(\tilde{S};Y),
\end{equation}
where $I$ denotes mutual information between these variables. 

A notion of information relevance comes from the second term, the \emph{relevance term}: with a positive coefficient $\beta$, this is encouraging summaries $\tilde{S}$ to contain information shared with $Y$. 
The first term, or \emph{pruning term}, ensures that irrelevant information is discarded. By minimizing the mutual information between summary $\tilde{S}$ and source $S$, any information about the source that is not credited by the \emph{relevance term} is thrown away.
The statistical structure of IB makes this compressive by forcing the summary to only contain information shared with the source.\footnote{In IB, this is a strict statistical relationship.}

In sum, IB relies on 3 principles: 
\begin{enumerate}
    \item Encouraging relevant information with a \emph{relevance term}.
    \item Discouraging extra information with a \emph{pruning term}.
    \item Strictly summarizing the source.
\end{enumerate}

To clarify the difference from a reconstructive loss, suppose there is irrelevant information in $S$ (i.e. unrelated to relevance variable $Y$), call this $Z$. With the IB objective (eq \ref{eq:bottlex_loss}), there is no benefit to keeping any information from $Z$, which strictly makes the first term worse (more mutual information between source and summary) and does not affect the second ($Z$ is unrelated to $Y$). In contrast, because $Z$ contains information about $S$, including it in $\tilde{S}$ could easily benefit the reconstructive loss (eq. \ref{eq:reconstructive}) despite being irrelevant.

As a relevance variable we will use the sentence following the source in the document in which it occurs.
This choice is motivated by linguistic cohesion, in which we expect more broadly relevant information to be common between consecutive sentences, while less relevant information and details are often not carried forward.

We use these principles to derive two methods for sentence summarization.
Our first method (\S \ref{sec:bottleEx}) enforces strict summarization through being extractive. 
Additionally, it does not require any training, so can be applied directly without the availability of domain-specific data.
The second method (\S \ref{sec:bottleSelf}) generalizes IB-based summarization to \emph{abstractive} summarization that can be trained on large unsupervised datasets, learning an explicit summarization function $p(\tilde{s}|s)$ over a distribution of inputs.

\section{Unsupervised Extractive Summarization} 
\label{sec:bottleEx}

We now use the Information Bottleneck principle to propose \bottleEx, an unsupervised extractive approach to sentence summarization. 
Our approach does not require any training; only a pretrained language model is required to satisfy the IB principles of (\ref{sec:IB}), and the stronger the language model, the stronger our approach will be.
In section \ref{sec:experiments} we demonstrate the effectiveness of this method using GPT-2, the pretrained language model of \citet{radford2019gpt2}. \footnote{We use the originally released ``small'' 117M parameter version.}

\subsection{IB for Extractive Summarization}
Here, we take advantage of the natural parallel between the Information Bottleneck and summarization developed in section \ref{sec:IB}. Working from the 3 IB principles stated there, we derive a set of actionable principles for a concrete sentence summarization method. 

We approach the task of summarizing a single sentence $s$ using the following sentence $s_{next}$ as the relevance variable. The method will be a deterministic function mapping $s$ to the summary $\tilde{s}$, so instead of learning a distribution over summaries, we take $p(\tilde{s} | s) = 1$ for the summary we arrive at. 
Our goal is then to optimize the IB equation (Eq \ref{eq:IB}) for a single example rather a distribution of inputs (as in the original IB method).

In this setting, to minimize equation \ref{eq:IB} we can equivalently minimize: 
\begin{equation} 
 - \log p(\tilde{s})  - \beta_1 p(s_{next} | \tilde{s}) p(\tilde{s}) \log p(s_{next} | \tilde{s}),
\label{eq:bottlex_loss}
\end{equation}
where coefficient $\beta_1 > 0$ controls the trade-off between keeping relevant information and pruning.
See appendix \ref{app:A} for the derivation of this equation.
Similar to eq \ref{eq:IB}, the first term encourages pruning, while the second encourages information about the relevance variable, $s_{next}$.
Both unique values in eq \ref{eq:bottlex_loss} ($p(\tilde{s})$ and $p(s_{next} | \tilde{s})$) can be estimated directly by a pretrained language model, a result of the summary being natural language as well as our choice of relevance variable. This will give us a direct path to enforcing IB principles 1 and 2 from section \ref{sec:IB}.

To interpret principle 3 for text, we consider what attributes are important to strict textual summarization. Simply, a strict textual summary should be shorter than the source, while agreeing semantically. The first condition is straightforward but the second is currently infeasible to ensure with automatic systems, and so we instead enforce extractive summarization to ensure the first and encourage the second.

Without a supervised validation set, there is no clear way to select a value for $\beta_1$ in Eq \ref{eq:bottlex_loss} and so no way to optimize this directly. Instead, we opt to ensure both terms improve as our method proceeds. Thus, we are not comparing the pruning and relevance terms directly (only ensuring mutual progress), and so we optimize simpler quantities monotonic in the two terms instead: $p(\tilde{s})$ for pruning and $p(y|\tilde{s})$ for relevance. 

We perform extractive summarization by iteratively deleting words or phrases, starting with the original sentence.
At each elimination step, we only consider candidate deletions which decrease the value of the pruning term, i.e., increase the language model score of the candidate summary. This ensures progress on the pruning term,
and also enforces the notion that word deletion should reduce the information content of the summary. 
The relevance term is optimized through only expanding candidates that have the highest relevance scores at each iteration, and picking the candidate with the highest relevance score as final summary.

Altogether, this gives 3 principles for extractive summarization with IB. 
\begin{enumerate}
    \item Maximize \emph{relevance term} by maximizing $p(s_{next}|\tilde{s})$.
    \item Prune information and enforce compression by bounding: $p(\tilde{s}_{i+1}) > p(\tilde{s}_i)$.
    \item Enforce strict summarization by extractive word elimination.
\end{enumerate}

\subsection{Method} 
\label{subsec:bottleExmethod}

\begin{algorithm}
\caption{\bottleEx{} method}\label{alg:bottleEx}
\begin{algorithmic}[1]
\Require sentence $s$ and context $s_{next}$ 
\State $C \gets \{s\}$ \Comment{set of summary candidates}
\For{$l$ in $length(s) ... 1$}
\State $C_l \gets \{s' \in C | len(s') = l\}$
\State sort $C_l$ descending by $p(s_{next}|s')$ 
\For{$s'$ in $C_l[1 \! : \! k]$} \label{line:topk}
\State $l' \gets length(s')$
\For{$j$ in $1...m$}  
\For{$i$ in $1...(l'-j)$}
\State $s'' \gets s'[1 \! : \! i \! - \! 1] \circ s'[i \! + \! j \! : \! l']$ \label{line:deletion}
\If {$p(s'') > p(s')$}
\State $C \gets C + \{s''\}$
\EndIf
\EndFor
\EndFor
\EndFor
\EndFor
\State \textbf{return} $\underset{s' \in C}{\argmax}\  p(s_{next}|s')$ \label{line:argmax}
\end{algorithmic}
\end{algorithm}

We turn these principles into a concrete method which iteratively produces summaries of decreasing length by deleting consecutive words in candidate summaries (Algorithm \ref{alg:bottleEx}). 
The relevance term is optimized in two ways: first, only the top-scoring summaries of each length are used to generate new, shorter summaries (line \ref{line:topk}). 
Second, the final summary is chosen explicitly by this measure (line \ref{line:argmax}).

In order to satisfy the second condition, each candidate must contain less self-information (i.e., have higher probability) than the candidate that derives it. This ensures that each deletion (line \ref{line:deletion}) strictly removes information. 
The third condition, strict extractiveness, is satisfied per definition. 

The algorithm has two parameters: $m$ is the max number of consecutive words to delete when producing new summary candidates (line \ref{line:deletion}), and $k$ is the number of candidates at each length used to generate shorter candidates by deletion (line \ref{line:topk}).

\section{Abstractive Summarization with Extractive Self-Supervision} \label{sec:bottleSelf}

Next, we extend the unsupervised summarization of \bottleEx{} to abstractive summarization with \bottleSelf{}, based on a straightforward technique for self-supervision. 
Simply, a large corpus of unsupervised summaries is generated with \bottleEx{} using a strong language model, then the same language model is tuned to produce summaries from source sentences on that dataset.

The conceptual goal of \bottleSelf{} is to use \bottleEx{} as a guide to 
learn the notion of information relevance as expressed through IB, but in a way that (a) removes the restriction of extractiveness, to produce more natural outputs and (b) learns an explicit compression function not requiring a next sentence for decoding.

\subsection{Extractive Dataset}
\label{subsec:bottleselfdata}

The first step of \bottleSelf{} is to produce a large-scale dataset for self-supervision using the \bottleEx{} method set out in \S \ref{subsec:bottleExmethod}.
The only requirement for the input corpus is that next sentences need to be available. 

In our experiments, we generate a corpus of 100,000 sentence-summary pairs with \bottleEx{}, using the same parameter settings as in section \ref{sec:bottleEx}. 
The resulting summaries have an average compression ratio (by character length) of approximately 0.55.

\subsection{Abstractive Fine-tuning}\label{subsec:bottleselfmodel}
The second step of \bottleSelf{} is fine-tuning the language model on its extractive summary dataset. 
The tuning data is formed by concatenating source sentences with generated summaries, separated by a delimiter and followed by an end token.
The model (GPT-2) is fine-tuned with a simple language modeling objective over the full sequence. %

As a delimiter, we use \texttt{ TL;DR: }, following \citet{radford2019gpt2} who found that this induces summarization behavior in GPT-2 even without tuning. We use a tuning procedure closely related to \newcite{radford2018improving}, training for 10 epochs. 
We take the trained model weights that minimize loss on a held-out set of 7000 extractive summaries. 

To generate from this model, we use a standard beam search decoder, keeping the top candidates at each iteration. Unless otherwise specified, assume we use a beam size of 5. We restrict produced summaries to be at least 5 tokens long, and no longer than the source sentence. 

\section{Experiments} \label{sec:experiments}
We evaluate our \bottleSum{ } methods using both automatic metrics and human evaluation. We find our methods dominant over a range of baselines in both categories.

\subsection{Setup}
We evaluate our methods and baselines using automatic ROUGE metrics (1,2,L) on the \textsc{DUC}-2003 and \textsc{DUC}-2004 datasets \cite{over2007duc}, similar to the evaluation used by \citet{SEQ3}. \textsc{DUC}-2003 and \textsc{DUC}-2004 consist of 624 and 500 sentence-summary pairs respectively. Sentences are taken from newstext, and each summary consists of 4 human-written reference summaries capped at 75 bytes. We recover next-sentences from \textsc{DUC} articles for \bottleEx{}. 

We also employ human evaluation as a point of comparison between models. This is both to combat known issues with ROUGE metrics \cite{schluter2017limits} and to experiment beyond limited supervised domains. Studying unsupervised methods allows for comparison over a much wider range of data where training summary pairs are not available, which we take advantage of here by summarizing sentences from the non-anonymized CNN corpus \cite{hermann2015cnn, nallapati2016abstractive, See17pointer-summarization}. 

We use Amazon Mechanical Turk (AMT) for human evaluation, summarizing on 100 sentences sampled from a held out set. Evaluation between systems is primarily done as a pairwise comparison between \bottleSum{} models and baselines, over 3 attributes: coherence, conciseness, and agreement with the input. AMT workers are then asked to make a final judgement of which summary has higher overall quality. Each comparison is done by 3 different workers. 
Results are aggregated across workers and examples.

\subsection{Models}

In both experiments, \bottleEx{} is executed as described in section \ref{subsec:bottleExmethod}. In experiments on \textsc{DUC} datasets, next-sentences are recovered from original news sources, while we limit test sentences in the CNN dataset to those with an available next-sentence (this includes over 95\% of sentences). We set parameter $k=1$ (i.e. expand a single candidate at each step) with up to $m=3$ consecutive words deleted per expansion. 
GPT-2 (small) is used as the method's pretrained language model, with no task-specific tuning. To clarify, the only difference between how \bottleEx{} runs on the datasets tested here is the input sentences; no data-specific learning is required.

As with \bottleEx{}, we use GPT-2 (small) as the base for \bottleSelf{}. To produce source-summary pairs for self supervision, we generate over 100,000 summaries using \bottleEx{} with the parameters above, on both the Gigaword sentence dataset (for automatic evaluation) and CNN training set (for human evaluation). \bottleSelf{} is tuned on the respective set for 10 epoch with a procedure similar to \citet{radford2019gpt2}. When generating summaries, \bottleSelf{} uses beam-search with beam size of 5, and outputs constrained to be at least 5 tokens long. 

We include a related model, \reconEx{} as a simple autoencoding baseline comparable in setup to \bottleEx{}. \reconEx{} follows the procedure of \bottleEx{}, but replaces the next-sentence with the source sentence. This aims to take advantage of the tendency of language models to semantically repeat in to substitute the Information Bottleneck objective in \bottleEx{} with a reconstruction-inspired loss. While this is not a perfect autoencoder by any means, we include it to probe the role of the next-sentence in the success of \bottleEx{}, particularly compared to a reconstructive method. As \reconEx{} tends to have a best reconstructive loss by retaining the entire source as its summary, we constrain its length to be as close as possible to the \bottleEx{} summary for the same sentence. 

As an unsupervised neural baseline, we include $\text{SEQ}^3$ \cite{SEQ3}, which is trained with an autoencoding objective paired with a topic loss and language model prior loss. $\text{SEQ}^3$ had the highest comparable unsupervised results on the \textsc{DUC} datasets that we are aware of, which we cite directly. For human evaluation, we retrained the model with released code on the training portion of the CNN corpus. 

We use the ABS model of \citet{rush2015namas} as a baseline for automatic and human evaluation. For automatic evaluation, this model is the best published supervised result we are aware of on the \textsc{DUC-2003} dataset, and we include it as a point of reference for the gap between supervised and unsupervised performance. We cite their results directly. For human evaluation, this model demonstrates the performance gap for out-of-domain summarization. Specifically, it requires supervision (unavailable for the CNN dataset), and so we use the model as originally trained on the Gigaword sentence dataset. This constitutes a significant domain-shift from the first-sentences of articles with limited vocabulary to arbitrary article sentences with diverse vocabulary.

We include the result of \citet{li-etal-2017-deep} on \textsc{DUC-2004}, who achieved the best supervised performance we are aware of. This is intended as a point of reference for supervised performance.

Finally, for automatic metrics we include common baseline \textsc{Prefix}, the first 75 bytes of the source sentence. To take into account lack of strict length constraints and possible bias of ROUGE towards longer sequences, we include \textsc{Input}, the full input sentence. Because our model is extractive, we know its outputs will be no longer than the input, but may exceed the length of other methods/baselines.

\subsection{Results}

\begin{figure*}
    \centering
    \includegraphics[origin=c,width=480pt,trim={0 10.7cm 3cm 0},clip]{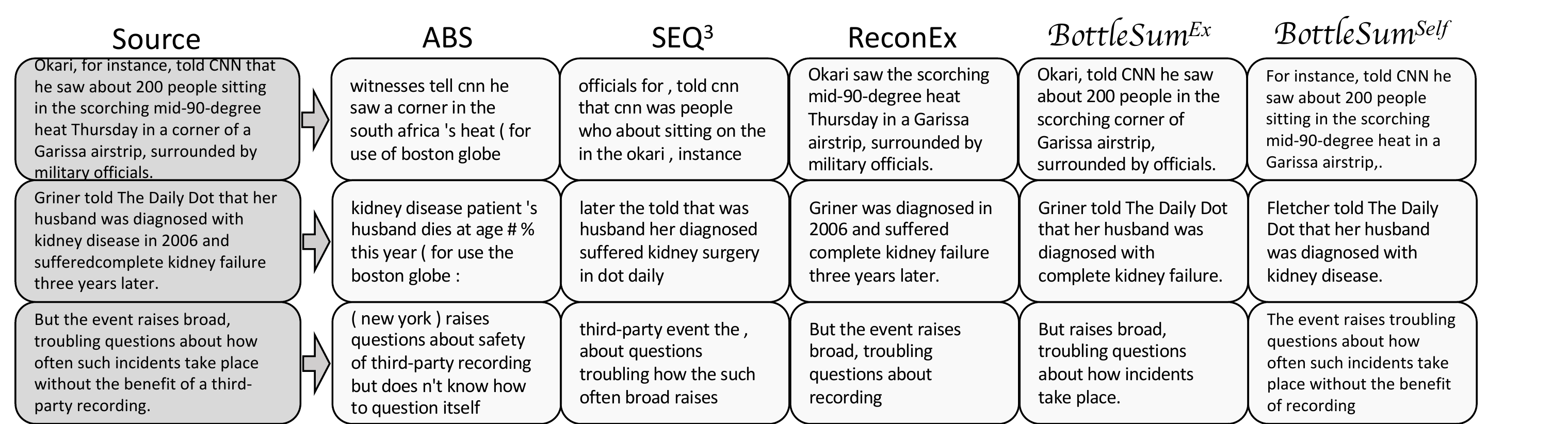}
    \caption{Representative example generations from the summarization systems compared}
    \label{fig:AEvIB-examples}
\end{figure*}
\begin{table}[t]
\small
    \centering
    \begin{tabular}{lccc}
     \toprule
 Method & R-1 & R-2 & R-L \\ 
 \midrule
\rowcolor[gray]{0.95} Supervised & \ccell{} & \ccell{} & \ccell{} \\
  ABS & 28.18 & 8.49 & 23.81\\ %
  \newcite{li-etal-2017-deep}& 31.79 & 10.75 & 27.48 \\
    \midrule
\rowcolor[gray]{0.95} Unsupervised & \ccell{} & \ccell{} & \ccell{} \\
    \textsc{Prefix} & 20.91 & 5.52 & 18.20\\ 
    \textsc{Input} & 22.18 & \textbf{6.30} & 19.33 \\
    $\text{SEQ}^3$ & 22.13 & 6.18 & 19.3 \\
    \reconEx{}  & 21.97 &	5.70 &	18.81 \\
    \midrule
    \bottleEx{}  & \textbf{22.85} &	5.71 &	\textbf{19.87} \\
    \bottleSelf{} & 22.30 &	5.84 &	19.60 \\
     \bottomrule
    \end{tabular}
    \caption{Averaged ROUGE on the DUC-2004 dataset} %
    \label{tab:duc2004}
\end{table}

\begin{table}[t]
\small
    \centering
    \begin{tabular}{lccc}
     \toprule
 Method & R-1 & R-2 & R-L \\ 
 \midrule
\rowcolor[gray]{0.95} Supervised & \ccell{} & \ccell{} & \ccell{} \\
   ABS & 28.48 & 8.91 & 23.97\\ 
    \midrule
\rowcolor[gray]{0.95} Unsupervised & \ccell{} & \ccell{} & \ccell{} \\
     \textsc{Prefix} & 21.14 & \textbf{6.35} & 18.74\\ 
     \textsc{Input} & 20.83 & 6.15 & 18.44 \\
    $\text{SEQ}^3$ & 20.90 & 6.08 & 18.55 \\ 
    \reconEx{} & 21.11 & 5.77 & 18.33 \\
    \midrule
    \bottleEx{}  & \textbf{21.80} & 5.63 & \textbf{19.19} \\ 
    \bottleSelf{} & 21.54 &	5.93 &	18.96 \\
     \bottomrule
    \end{tabular}
    \caption{Averaged ROUGE on the DUC-2003 dataset}  %
    \label{tab:duc2003}
\end{table}

\begin{table*}[t]
\small
    \centering
    \begin{tabular}{lc|ccc|ccc}
    \toprule
    \textbf{Models} &&& \textbf{Attributes} &&& \textbf{Overall} \\
    \midrule
  Model & Comparison & coherence & conciseness & agreement & better & equal & worse \\ 
    \midrule
    \bottleEx{} vs. & ABS & +0.45 & +0.48 & +0.52 & 60\% & 31\% & 9\%\\
    & $\text{SEQ}^3$ & +0.61 & +0.57 & +0.56 & 61\%& 34\%& 5\%\\
    & \reconEx{} & -0.05 & +0.01 & -0.05 & 37\% & 22\% & 41\% \\
  \midrule
    \bottleSelf{} vs. & ABS & +0.47 &  +0.39 & +0.48 & 62\% & 26\% & 12\%\\
    & $\text{SEQ}^3$ & +0.56 & +0.45 & +0.53 & 65\% & 26\% & 9\% \\
    & \reconEx{} & +0.11 & -0.05 & +0.09 & 47\% & 14\% & 39\% \\
    & \bottleEx{} & +0.14 & +0.06 & +0.11 & 43 \% & 27\%& 30\% \\
    \bottomrule
    \end{tabular}
    \caption{Human evaluation on 100 CNN test sentences (pairwise comparison of model outputs). 
     Attribute scores are averaged over a scale of 1 (better), 0 (equal) and -1 (worse). We also report the overall preferences as percentages. }
    
    \label{tab:human}
\end{table*}

In automatic evaluation, we find \bottleEx{} achieves the highest R-1 and R-L scores for unsupervised summarization on both datasets. This is promising in terms of the effectiveness of the Information Bottleneck (IB) as a framework. \bottleSelf{} achieves the second highest scores in both of these categories, further suggesting that the tuning process used here is able to capture some of this benefit. The superiority of \bottleEx{} suggests possible benefit to having access to a relevance variable (next-sentence) to the effectiveness of IB on these datasets. 

The R-2 scores for \bottleEx{} on both benchmark sets were lower than baselines, possibly due to a lack of fluency in the outputs of the extractive approach used. \textsc{Prefix} and \textsc{Input} both copy human text directly and so should be highly fluent, while \citet{rush2015namas} and \citet{SEQ3} have the benefit of abstractive summarization, which is less restrictive in word order. Further, the fact that \bottleSelf{} is abstractive and surpasses R-2 scores of both new extractive methods tested here (\bottleEx{}, \reconEx{}) supports this idea. \reconEx{}, also extractive, has similar R-2 scores to \bottleEx{}.

The performance of \reconEx{}, our simple reconstructive baseline, is mixed. It does succeed to some extent (e.g. surpassing R-1 for all other baselines but \textsc{Prefix} on \textsc{Duc-2003}) but not as consistently as either \bottleSum{} method. This suggests that while some benefit may come from the extractive process of \bottleEx{} alone (which \reconEx{} shares), there is significant benefit to using a strong relevance variable (specifically in contrast to a reconstructive loss).

Next, we consider model results on human evaluation. \bottleSelf{} and \bottleEx{} both show reliably stronger performance compared to models from related work (ABS and $\text{SEQ}^3$ in Table \ref{tab:human}). While \bottleSelf{} seems superior to \reconEx{} other than in conciseness (in accordance with their compression ratios in Table \ref{tab:abstractiveness}), \bottleEx{} appears roughly comparable to \reconEx{} and slightly inferior to \bottleSelf{}. 

The inversion of dominance between \bottleEx{} and \bottleSelf{} on automatic and human evaluation may cast light on competing advantages. \bottleEx{} captures reference summaries more effectively, while \bottleSelf{}, through a combination of abstractivness and learning a cohesive underlying mechanism of summarization, writes more favorable summaries for a human audience. Further analysis and accounting for known limitations of ROUGE metrics may clarify these competing advantages.

In comparing these models, there are also practical considerations (summarized in table \ref{tab:comparison}). ABS can be quite effective, but requires learning on a large supervised training set (as demonstrated by its poor out-of-domain performance in Table \ref{tab:human}). While $\text{SEQ}^3$ is unsupervised, it still needs extensive training on a large corpus of in-domain text. \bottleEx{}, whose outputs were preferred over both by humans, requires neither of these. Given a strong pretrained language model (GPT-2 small is used here) it only requires a source and next-sentence to summarize. \bottleSelf{} requires in-domain text for self-supervision, but its superior performance by human evaluation and summarization without next-sentence are clear advantages. Further, its beam-search decoding is more computationally efficient than \bottleEx{}, which requires evaluating conditional next-sentence perplexity over a large grid of extractive summary candidates.

Another difference from \bottleEx{} is the ability of \bottleSelf{} to be abstractivene (Table \ref{tab:abstractiveness}). Other baselines have a higher degree of abstractiveness than \bottleSelf{}, but this can be misleading. Consider the examples in figure \ref{fig:AEvIB-examples}. While many of the phrases introduced by other models are technically abstractive, they are often off-topic and confusing.

This hints at an advantage of \bottleSum{} methods. In only requiring the base model to be a (tunable) language model, they are architecture-agnostic and can incorporate as powerful a language model as is available. Here, incorporating GPT-2 (small) carries benefits like strong pretrained weights and robust vocabulary handling by byte pair encoding, allowing them to process the diverse language of the non-anonymized CNN corpus with ease. The specific benefits of GPT-2 are less central, however; any such language model could be used for \bottleEx{} immediately, and \bottleSelf{} with some tuning. This is in contrast architecture-specific models like ABS and $\text{SEQ}^3$, which would require significant restructuring to fully incorporate a new model.

As a first work to study the Information Bottleneck principle for unsupervised summarization, our results suggest this is a promising direction for the field. It yielded two methods with unique performance benefits (Table \ref{tab:duc2004}, \ref{tab:duc2003}, \ref{tab:human}) and practical advantages (table \ref{tab:comparison}). We believe this concept warrants further exploration in future work. 

\begin{table}[t]
\small
\centering
\begin{tabular}{lll}  
\toprule
Model                  & Abstractive & Compression \\
                  & Tokens \% & Ratio \% \\ 
\midrule
\bottleEx{}               & -                    & 51   \\
\reconEx{}               & -                    & 52   \\
\bottleSelf{}              & 5.8           & 56  \\         
$\text{SEQ}^3$ & 12.6                   & 58 \\  
ABS                  & 60.4    & 64 \\          
\bottomrule
\end{tabular}
\caption{Abstractiveness and compression of CNN summaries. Abstractiveness is omitted for strictly extractive approaches} 

\label{tab:abstractiveness}
\end{table}

\begin{table*}[t]
    \centering
    \small
    \begin{tabular}{p{1.5cm}|p{2.1cm}p{2.4cm}p{2.4cm}p{5.9cm}}
    \toprule
    \textbf{Model Name} & \textbf{Model architecture} & \textbf{Data for training} & \textbf{Data for summarizing} & \textbf{When to use} \\
    \midrule
    ABS \cite{rush2015namas} & Seq2Seq & Large scale, paired source-summaries (supervised) & source sentence & Large scale supervised training set available \\ 
    \midrule
    $\text{SEQ}^3$ \cite{SEQ3} & Seq2Seq2Seq
    & Large scale, unsupervised source sentences & source sentence & Large scale unsupervised (no summaries) data available, \emph{without} next-sentences  \\
    \midrule
    \bottleEx{} & Pre-trained LMs & No training data needed & source sentence and next-sentence
    & No training data available, and next-sentences \emph{are} available for sentences to summarize. \\
    \midrule
    \bottleSelf{} & Pre-trained LMs \newline
    (fined-tuned on data from \bottleEx{})
    & Large scale, unsupervised source sentences with next sentences & source sentence & Large scale unsupervised (no summaries) data available, \emph{with} next-sentences and/or no next-sentences available for sentences to summarize \\
    \bottomrule
    \end{tabular}
    \caption{Comparison of sentence summarization methods.} 
    \label{tab:comparison}
\end{table*}

\section{Related Work}
\subsection{Sentence Compression and Summarization}

\citet{rush2015namas} first proposed abstractive sentence compression with neural sequence to sequence models, 
trained on a large corpus of headlines with the first sentences of articles as supervision.
This followed early work on approaching headline generation as statistical machine translation \cite{banko-etal-2000-headline}. 
Subsequently, recurrent neural networks with pointer-generator decoders became standard for this task, and focus shifted to the document-level \cite{nallapati2016abstractive,See17pointer-summarization}.

Pointer-based neural models have also been proposed for extractive summarization \cite{cheng-lapata-2016-neural}. 
The main limitations of this approach are that the training data is constructed heuristically, covering a specific type of sentence summarization (headline generation).
Thus, these supervised models do not generalize well to other kinds of sentence summarization or domains. In contrast, our method is applicable to any domain for which example inputs are available in context.

\subsection{Unsupervised Summarization}

\citet{miao-blunsom-2016-language} framed sentence compression as an autoencoder problem, where the compressed sentence is a latent variable from which the input sentence is reconstructed.
They proposed extractive and pointer-generator models, regularizing the autoencoder with a language model to encourage compression and optimizing with the REINFORCE algorithm.
While their extractive model does not require supervision, results are only reported for semi-supervised training, using less supervised data than purely supervised training.
\citet{fevry-phang-2018-unsupervised} applied denoising autoencoders to fully unsupervised summarization, while \citet{wang-lee-2018-learning} proposed an autoencoder with a discriminator for distinguising well-formed and ill-formed compressions in a Generative Adversarial Network (GAN) setting, instead of using a language model. 
However, their discriminator was trained using  unpaired summaries, so while they beat purely unsupervised approaches like ours their results are not directly comparable.
Recently \citet{SEQ3} proposed a differentiable autoencoder using a gumbel-softmax to represent the distribution over summaries.
The model is trained with a straight-through estimator as an alternative to reinforcement learning, obtaining better results on unsupervised summarization.
All of these approaches have in common autoencoder-based training, which we argue does not naturally capture information relevance for summarization. 

Recently, \citet{zhou2019simple} introduced a promising method for summarization using contextual matching with pretrained language models. While contextual matching requires pretrained language models to generate contextual vectors, \bottleSum{} methods do not have specific architectural constraints. Also, like \citet{wang-lee-2018-learning} it trains with unpaired summaries and so is not directly comparable to us. 

\subsection{Mutual Information for Unsupervised Learning}

We take inspiration from an exciting direction leveraging mutual information for unsupervised learning. Recent work in this area has seen success in natural language tasks \cite{mcallester2018information, Oord2018RepresentationLW}, as well as computer vision \cite{Bachman2019LearningRB, hjelm2019learning} by finding novel ways to measure and optimize mutual information. Within this context, our work is a further example suggesting mutual information is an important element stimulating progress in unsupervised learning and modelling.

\section{Conclusion}
We have presented \bottleEx{}, an unsupervised extractived approach to sentence summarization, and extended this to \bottleSelf{}, a self-supervised abstractive approach. \bottleEx{}, which can be applied without any training, achieves competitive performance on automatic and human evaluations, compared to unsupervised baselines.
\bottleSelf{}, trained on a new domain, obtains stronger performance by human evaluation than unsupervised baselines as well as \bottleEx{}.
Our results show that the Information Bottleneck principle, by encoding a more appropriate notion of relevance than autoencoders, offers a promising direction for progress on unsupervised summarization. 

\section{Acknowledgments}
We thank anonymous reviewers for many helpful comments. 
This research is supported in part by the Natural Sciences and Engineering Research Council of Canada (NSERC) (funding reference number 401233309), NSF (IIS-1524371),  DARPA CwC through ARO (W911NF15-1-0543), Darpa MCS program N66001-19-2-4031 through NIWC Pacific (N66001-19-2-4031), Samsung AI Research, and Allen Institute for AI.



\bibliography{emnlp-ijcnlp-2019}
\bibliographystyle{acl_natbib}
\clearpage
\section{Appendix}
\appendix
\section{Derivation of \bottleEx{} Loss} \label{app:A}
This is a derivation of the loss equation (\ref{eq:bottlex_loss}) used in section \ref{sec:IB}, starting with the Information Bottleneck (IB) loss given in eq \ref{eq:IB}:
\begin{equation}
    I(\tilde{S};S) - \beta I(\tilde{S};Y),
\end{equation}
The goal here is to consider how to interpret this equation when we only have access to single values of source $S$ and relevance variabel $Y$ at a time. In the original IB formulation, a distribution $p(\tilde{S}|S)$ to go from sources to summaries is learned by optimizing this expression across the distribution of source-target pairs $(s,y)$. In the case of \bottleEx{}, the goal is to consider this expression on a case-by-case basis, not requiring training over a large distribution of pairs.

First, we consider an alternate form of the equation above:
\begin{equation} 
\begin{split}
&I(\tilde{S};S) - \beta I(\tilde{S};Y) \\
  = & \underset{S,\tilde{S}}{\mathbb{E}}[pmi(\tilde{S};S)] - \beta \underset{Y,\tilde{S}}{\mathbb{E}}[ pmi(\tilde{S};Y)]
\end{split}
\end{equation}

where $pmi(x,y) = \frac{p(x,y)}{p(x)p(y)}$ denotes pointwise mutual information. 
\begin{equation} 
\begin{split}
  = & \underset{S,\tilde{S}}{\mathbb{E}} \left [pmi(\tilde{S};S) \right ] - \beta \underset{Y,\tilde{S}}{\mathbb{E}} \left [ pmi(\tilde{S};Y) \right] \\
\end{split}
\end{equation}
As stated above, we want to consider this for only single values of $s$ and $y$ at a time, so for these values we can investigate the applicable terms of these expectations:
\begin{equation} 
\begin{split}
  = & \sum_{\tilde{s}} \left [ p(s,\tilde{s})pmi(\tilde{S};S)  - \beta p(y,\tilde{s})pmi(\tilde{S};Y) \right]
\end{split}
\end{equation}
This is the expression we are then aiming to optimize, as it covers all terms in the original IB objective that we have access to on a case-by-case basis. 

As in the original IB problem, we can think of learning a distribution $p(\tilde{s}|s)$. However, we are now only taking an expectation over $\tilde{S}$ and so we simply collapse all probability onto the setting of $\tilde{s}$ that optimizes this expression. Simply:
\begin{equation}
    p(\tilde{s}|s) = 1 \text{ for chosen summary, } 0 \text{ otherwise}
\end{equation}

This results in finding $\tilde{s}$ that optimizes:

\begin{equation} 
\begin{split}
   &  p(s,\tilde{s})pmi(\tilde{S};S)  - \beta p(y,\tilde{s})pmi(\tilde{S};Y) \\
  = &  p(s,\tilde{s})\text{log} \frac{p(s,\tilde{s})}{p(s)p(\tilde{s})}  - \beta p(y,\tilde{s})\text{log}\frac{p(y,\tilde{s})}{p(y)p(\tilde{s})} 
\end{split}
\end{equation}
Any terms that rely only on $s$ and $y$ will be constant and so can be collected into coefficients. As well, remembver that we set $p(\tilde{s}|s) = 1$. Doing rearranging:
\begin{equation} 
\begin{split}
  = &  p(\tilde{s}|s)p(s)\text{log} \frac{p(\tilde{s}|s)}{p(\tilde{s})}  - \beta p(y|\tilde{s})p(\tilde{s})\text{log}\frac{p(y|\tilde{s})}{p(y)} \\
    = &  c_1 \text{log} \frac{1}{p(\tilde{s})}  - \beta p(y|\tilde{s})p(\tilde{s})\text{log }p(y|\tilde{s})- c_2
\end{split}
\end{equation}
This is equivalent to optimizing:
\begin{equation} \label{eq:bottleEx_obj_appendix}
\begin{split}
 \text{log} \frac{1}{p(\tilde{s})}  - \beta_1 p(y|\tilde{s})p(\tilde{s})\text{log }p(y|\tilde{s})
\end{split}
\end{equation}
for some positively signed $\beta_1$. 


\end{document}